\documentclass[letterpaper, 10 pt, journal, twoside]{IEEEtran}

\usepackage{amsmath} 
\usepackage{xcolor}
\usepackage{hyperref}
\usepackage{cite}
\usepackage{graphicx}
\usepackage{amsmath}
\usepackage{bm}
\usepackage{multirow}
\usepackage{booktabs}
\usepackage{array}
\usepackage{pifont}%
\usepackage[symbol]{footmisc}
\newcommand{\cmark}{\ding{52}}%
\newcommand{\xmark}{\ding{56}}%

\usepackage{algorithm}
\usepackage{algorithmicx}
\usepackage[noend]{algpseudocode}

\title{\LARGE \bf
SOUS VIDE: Cooking Visual Drone Navigation Policies in a Gaussian Splatting Vacuum}

\author{JunEn Low, Maximilian Adang, Javier Yu, Keiko Nagami, and Mac Schwager

\thanks{Manuscript received: December 20, 2024; Revised: February 18, 2025; Accepted: March 7, 2025. This paper was recommended for publication by Editor Pascal Vasseur upon evaluation of the Associate Editor and Reviewers’ comments. This work was supported in part by DARPA grant HR001120C0107, ONR grant N00014-23-1-2354, and Lincoln Labs grant 7000603941. The second author was supported on an NDSEG fellowship. Toyota Research Institute provided funds to support this work.}
\thanks{The authors are with Stanford University, Stanford, CA 94404, USA (e-mail: jelow@stanford.edu, madang@stanford.edu, javieryu@stanford.edu, knagami@stanford.edu, schwager@stanford.edu)}%
\thanks{Digital Object Identifier (DOI): see top of this page.}
}

\begin{document}

\markboth{IEEE Robotics and Automation Letters. Preprint Version. Accepted March, 2025}{Low \MakeLowercase{\textit{et al.}}: SOUS VIDE: Cooking Visual Drone Navigation Policies in a Gaussian
Splatting Vacuum} 

\maketitle

\begin{abstract}
We propose a new simulator, training approach, and policy architecture, collectively called SOUS VIDE, for end-to-end visual drone navigation. Our trained policies exhibit zero-shot sim-to-real transfer with robust real-world performance using only onboard perception and computation. Our simulator, called FiGS, couples a computationally simple drone dynamics model with a high visual fidelity Gaussian Splatting scene reconstruction. FiGS can quickly simulate drone flights producing photorealistic images at up to 130 fps.  We use FiGS to collect 100k-300k image/state-action pairs from an expert MPC with privileged state and dynamics information, randomized over dynamics parameters and spatial disturbances. We then distill this expert MPC into an end-to-end visuomotor policy with a lightweight neural architecture, called SV-Net. SV-Net processes color image, optical flow and IMU data streams into low-level thrust and body rate commands at 20 Hz onboard a drone. Crucially, SV-Net includes a learned module for low-level control that adapts at runtime to variations in drone dynamics. In a campaign of 105 hardware experiments, we show SOUS VIDE policies to be robust to 30\% mass variations, 40 m/s wind gusts, 60\% changes in ambient brightness, shifting or removing objects from the scene, and people moving aggressively through the drone's visual field. Code, data, and experiment videos can be found on our project page: \url{https://stanfordmsl.github.io/SousVide/}.
\end{abstract}

\section{INTRODUCTION}
\label{sec:introduction}
\IEEEPARstart{L}{earned} visuomotor policies offer a compelling alternative to traditional drone navigation stacks by unifying perception and control into a streamlined framework. Unfortunately, training policies with human-like agility and collision avoidance requires a large corpus of visual and state data, making behavior cloning from human pilot demonstrations impractical. Simulation provides a promising alternative, but the sim-to-real gap has remained a persistent obstacle to real-world deployment. Recent work has demonstrated that in controlled environments and with carefully built digital twins in simulation, learned policies can achieve highly agile, superhuman performance \cite{loquercio2021learning,kaufmann2023champion,gelesRSS2024demonstrating}. However, this raises the question: can we train a visuomotor policy to navigate unstructured real-world environments with minimal human curation?

\begin{figure}[t]
  \centering
  \includegraphics[width=\columnwidth]{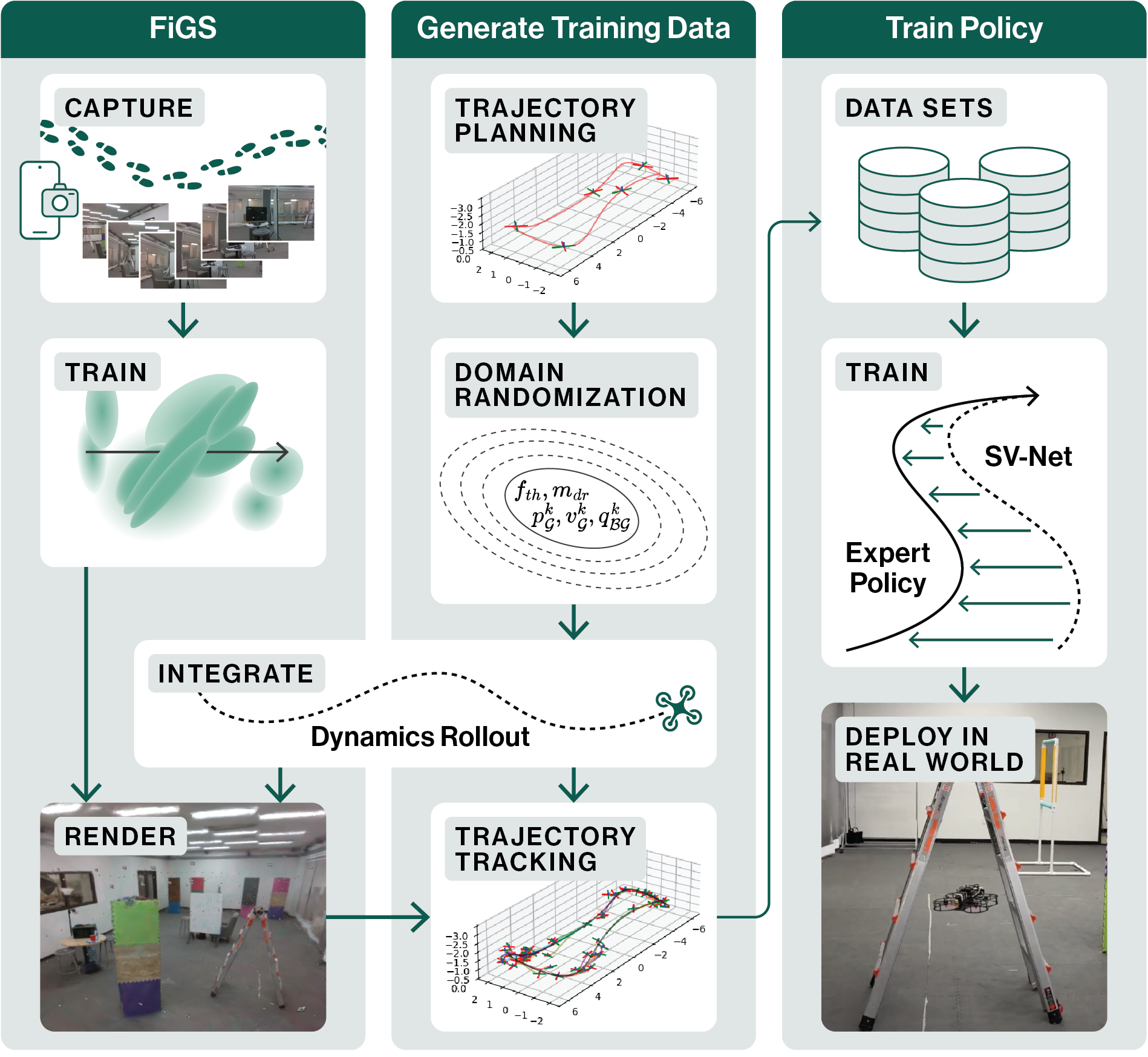}
  \vspace{-7.5mm}
  \caption{SOUS VIDE overview: We train our FiGS simulator from a hand held camera. We use FiGS to generate flight demonstrations (image/state-action pairs) from an MPC expert with privileged information randomized over dynamics parameters and positional disturbances. We use this data to train our policy, SV-Net, which operates solely with onboard observations.}
  \label{fig:pipeline_diagram}
  \vspace{-5.5mm}
\end{figure}

We address this challenge with FiGS (Flying in Gaussian Splats), a photorealistic drone simulator combining a Gaussian Splat (GSplat) \cite{kerbl20233d} scene model with a lightweight 10-dimensional drone dynamics model using thrust and body rate inputs—equivalent to the Acro mode used by expert human pilots. FiGS reconstructs scenes from video captures, like publicly available footage or smartphone recordings, processed with standard tools \cite{nerfstudio} and generates realistic image sequences and state estimation data for a drone at up to 130 fps on a standard GPU, all within an hour of acquiring video data. This contrasts with the current practice of approximating each real scene with a handcrafted simulation instance \cite{shah2018airsim,guerra2019flightgoggles,song2021flightmare,panerati2021learning,10556959,xu2023omnidrones} that can take weeks of laborious sim-to-real transfer to perfect.

Building on FiGS, we develop SOUS VIDE\footnote[2]{The name is inspired by the sous vide cooking technique that evenly cooks food in a vacuum-sealed bag under carefully controlled conditions.} (Scene Optimized Understanding via Synthesized Visual Inertial Data from Experts), a behavior cloning pipeline that produces a robust drone navigation policy capable of zero-shot sim-to-real transfer---trained entirely within simulation without real-world demonstrations or fine-tuning (Fig.~\ref{fig:pipeline_diagram}). Specifically, we use FiGS to generate 100k-300k image/state-action pairs from an expert MPC policy following a desired nominal trajectory within a GSplat. The MPC has access to the ground truth state in the simulator and is therefore able to demonstrate high-quality, collision-free trajectories. To obtain stable and robust flight, we randomize dynamics parameters and spatial perturbations and record the MPC's response. We use these expert demonstrations to train a student policy without privileged information (no lateral position data).

The learned policy produced by SOUS VIDE, SV-Net, is a novel architecture designed to process both images and observable state history while remaining efficient enough to run onboard the drone. The policy ingests images using a SqueezeNet \cite{iandola2016squeezenet}, re-trained on our data, and outputs a feature vector that is fused with observable data across several small Multi-Layer Perceptrons (MLPs) to produce low-level body-rate commands. Within these MLPs, we also implement a form of the Rapid Motor Adaptation (RMA) concept from \cite{kumar2021rma}, which takes a history of observable data from a sliding window and produces a latent code that captures evolving flight dynamics in real time. We find this RMA module is crucial for robust flight, adapting online to variations such as battery drain, rotor downwash effects, and wind gusts.

To summarize, we make the following contributions:
\begin{enumerate}
    \item \textbf{Flying in Gaussian Splats (FiGS)}: A simulator coupling GSplat scene models with drone dynamics for efficient and photorealistic visual-inertial data generation.
    \item \textbf{Scalable Visuomotor Policy Generation}: We use FiGS to generate large synthetic datasets to train visuomotor policies that transfer zero-shot to real-world flight.
    \item \textbf{SV-Net}: An onboard policy that fuses image and observable states to infer thrust and body rates while continuously adapting to changing flight conditions.
\end{enumerate}

We evaluate SOUS VIDE policies across 105 hardware flights in 4 different scenes, testing 9 different experimental conditions. We demonstrate our policy's robustness to 30\% mass variations, 40 m/s wind gusts, 60\% changes in ambient brightness, shifting or removing objects from the scene, and people moving aggressively through the drone's visual field.

The paper is organized as follows: Section \ref{sec:related_work} reviews related work, Section \ref{sec:figs} describes the FiGS simulator, Section \ref{sec:expert_data_synthesis} details the MPC-based data synthesis, and Section \ref{sec:sv-net} presents the SV-Net policy. Hardware experiments are in Section \ref{sec:experiments}, with conclusions, limitations, and future work in Section \ref{sec:conclusions}.

\section{Related Work} \label{sec:related_work}
\textit{Drone Polices Trained with GSplats:}  Learned representations like GSplats have proven effective in training visuomotor policies across many domains, from manipulation \cite{zhou2023nerf,qureshi2024splatsim} to bipedal locomotion \cite{haarnoja2024learning} to aerial robotics \cite{tagliabue2024tube,quach2024gaussian}. Closest to our approach, \cite{tagliabue2024tube} uses a learned representation and an MPC expert to train a trajectory-following policy but requires an initial sample of real-world expert flight demonstrations to be collected via motion capture. Additionally, its $45^\circ$ downward-facing camera focuses on ground features, missing the spatial information of a forward view needed for obstacle avoidance, as seen in drone racing \cite{kaufmann2023champion}. Meanwhile, \cite{quach2024gaussian}, treats the GSplat reconstruction as a background, relying instead on colored spheres as visual markers injected into both the simulation and real-world scenes as the basis for decision making. Moreover, it relies on velocity commands that encode only high-level approach and turn behaviors, delegating low-level control to a manufacturer-supplied autonomy stack. To the best of our knowledge, SOUS VIDE is the first method to leverage GSplats for generating low-level drone navigation policies for unstructured environments without assistive infrastructure or real-world expert flight data.

\textit{Training Drone Policies in Simulation:}
Simulators offer scalability and safety in collecting training data, but they introduce a sim-to-real gap, making it difficult to transfer policies to the real world. Many existing works use domain randomization \cite{tobin2017domain} to robustify policies, as we also do. For drones, another common strategy is to enhance the fidelity of the drone dynamics model with drag and other effects \cite{hoffmann2011precision,faessler2017differential,tal2020accurate}, while another is to improve rendering pipelines and graphics assets \cite{shah2018airsim,guerra2019flightgoggles,song2021flightmare,panerati2021learning,10556959,xu2023omnidrones}. However, none of these approaches can match the speed and visual fidelity of GSplat scene reconstructions. Another prevalent solution involves visual abstractions, such as depth maps \cite{loquercio2021learning,bhattacharya2024vision} or learned feature embeddings \cite{kaufmann2023champion}, which aim to distill visual information into a domain-invariant representation. However, this discards information encoded in raw pixel data that could otherwise improve task performance.

High-performance simulation-trained policies have been demonstrated for drone racing \cite{kaufmann2023champion,gelesRSS2024demonstrating}, marking an impressive technological achievement.  However, these methods blend real and simulation flight data, physics and learning-based models, and hand-engineered visual features cued into racing gates.  In contrast, our method can train a policy using video clips of the scene and can transfer zero-shot to the real-world with only minimal tuning of easily measurable parameters.

\textit{Rapid Motor Adaptation:} RMA, a technique originally developed for quadruped locomotion policies in \cite{kumar2021rma}, can be viewed as a pre-trained alternative to online parameter estimation \cite{loianno2016estimation} where an encoder is trained to take in a sensing history to produce a latent vector that captures runtime operating conditions (e.g., terrain for a quadruped, or flight dynamics for a drone). RMA has been adapted for drones in \cite{zhang2023learning,zhang2024learning} where they have been show to achieve stable flight with impressive robustness. However, they are not designed for visual navigation. Our lightweight RMA implementation is crucial for real-world robustness, addressing variations in both modeled and unmodeled drone dynamics.

\textit{Generalist Collision Avoidance Policies:} Some existing works train policies to steer a drone through environments not seen at training time, often focusing on a particular scene domain like forests \cite{ross2013learning}, office buildings \cite{sadeghi2016cad2rl}, or urban roadways \cite{loquercio2018dronet}.  Such policies have been trained both with Reinforcement Learning (RL) \cite{sadeghi2016cad2rl} and with Behavior Cloning (BC) \cite{ross2013learning,loquercio2018dronet}, using both simulated \cite{sadeghi2016cad2rl,bhattacharya2024vision} and real-world \cite{ross2013learning,gandhi2017learning,shah2023gnm,doshi2024scaling} data. Recent examples strive toward policies that can operate across different robot embodiments \cite{shah2023gnm,doshi2024scaling}. While impressive for their generality, they often treat the drone as a pseudo-static ground robot by using a finely tuned onboard Visual-Inertial Odometry (VIO) stack to constrain the dynamics to planar velocities and yaw. This fails to exploit the drone's full agility when navigating cluttered indoor environments with complex 3D trajectories. In contrast, SOUS VIDE directly commands thrust and body rates, mirroring the capabilities of expert human pilots.
\section{Flying in Gaussian Splats (FiGS)} \label{sec:figs}
FiGS, our lightweight GSplat-based flight simulator, consists of a GSplat model trained from video captures of the scene, within which a drone is simulated using a simplified 10-dimensional drone dynamics model.

\textit{Gaussian Splats:}
3D Gaussian Splatting \cite{kerbl20233d} is a learned representation approach that approximates the geometry and appearance of real-world scenes using a large collection of Gaussians—potentially millions—each parameterized by its position, covariance, color, and opacity. They leverage high-speed, projection-based differentiable rasterization and are trained from sparse RGB images by backpropagating through the rasterization to minimize photometric error. This approach enables photorealistic reconstructions and full-resolution renders at over 100 fps on a standard desktop GPU.

In this work, we generate GSplats from short video recordings (2-3 minutes) of scene walk-throughs with a handheld camera. From the video we extract a set of training images and use the open-source tool Nerfstudio \cite{nerfstudio,Ye_gsplat} to train the GSplat model. The resulting model $\mathcal{GS}_{\phi}$, where $\phi$ are its parameters, can render photorealistic images from a virtual camera placed at any pose within the region covered by the training images. Given a camera pose $(\bm{p}, \bm{q})$, where $\bm{p}$ represents the position and $\bm{q}$ the orientation in quaternion form, the rendered image is given by $\bm{I}= \mathcal{GS}_{\phi}(\bm{p}, \bf{q})$. To obtain metric scale and align the GSplat frame to a known global frame in the scene, we start the video recording with an ArUco tag marker in frame.

\textit{Drone Dynamics Model:}
Our model operates in the world, body, and camera frames ($\mathcal{W}$, $\mathcal{B}$, $\mathcal{C}$) and uses a 10-dimensional semi-kinematic state vector,  
$\bm{x} = \begin{bmatrix} \bm{p}_\mathcal{W},\bm{v}_\mathcal{W}, \bm{q}_\mathcal{BW} \end{bmatrix}^T$, representing position $\bm{p}_\mathcal{W} = (p_x, p_y, p_z)$, velocity $\bm{v}_\mathcal{W} = (v_x, v_y, v_z)$, and orientation $\bm{q}_\mathcal{BW} = (q_x, q_y, q_z, q_w)$. The control inputs,  
$\bm{u} = \begin{bmatrix} f_{th}, \bm{\omega}_\mathcal{B}\end{bmatrix}^T$, include normalized thrust $f_{th}$ and angular velocity $\bm{\omega}_\mathcal{B} = (\omega_x, \omega_y, \omega_z)$. This produces model dynamics
\begin{equation}\label{eq:equations_of_motion}
\begin{split}
\bm{\dot{p}}_\mathcal{W} &= \bm{v}_\mathcal{W},\\[-1ex]
\bm{\dot{v}}_\mathcal{W} &= g\bm{z}_\mathcal{W} - k_{th}\frac{f_{th}}{m_{dr}} \bm{z}_\mathcal{B} \\[-1ex]
\bm{\dot{q}}_\mathcal{BW} &= \frac{1}{2} \bm{W}(\bm{\omega}_\mathcal{B})\bm{q}_\mathcal{BW},
\end{split},
\end{equation}
where $g$ is gravitational acceleration, $\bm{W}(\bm{\omega}_\mathcal{B})$ is the quaternion multiplication matrix, and $\bm{z}_\mathcal{W}$, $\bm{z}_\mathcal{B}$ are the z-axis unit vectors of the world and body frames. The thrust coefficient and mass, $(k_{th}, m_{dr})$, are stored in the drone parameter vector $\bm{\theta}$.

Thrust and body rate commands are the standard low-level input for most flight controllers \cite{faessler2017differential,loquercio2021learning,kaufmann2023champion,gelesRSS2024demonstrating}, providing robust tracking through high-rate gyroscope feedback. This choice also enhances platform agnosticism in a cost-effective manner and is widely favored by expert human pilots. Moreover, for our use case, it offers the significant advantage of omitting the rotational acceleration equations (Euler's equations) in (\ref{eq:equations_of_motion}).

We forward integrate these equations of motion using ACADOS \cite{Verschueren2021}, a highly efficient trajectory optimizer that provides direct access to its dynamics update function, to obtain the state trajectory $\mathbf{X} = \{\bm{x}_0, \dots, \bm{x}_K \}$ and input trajectory $\mathbf{U} = \{\bm{u}_0, \dots, \bm{u}_{K-1} \}$, where $K$ denotes the number of discrete time steps. Applying the body-camera transform $T_\mathcal{C}^\mathcal{B}$ to the pose variables within $\mathbf{X}$, we can render the image sequence $\bm{\mathcal{I}} = \{\bm{I}_0, \dots, \bm{I}_K \}$ as seen by the onboard camera from the GSplat. This data can be used in an RL or BC framework, and can supervise the training of either state-feedback or image-feedback policies. For SOUS VIDE, we use an BC framework for image/state feedback, which we will describe next.

\section{MPC Expert and Data Synthesis}
\label{sec:expert_data_synthesis}
SOUS VIDE generates visuomotor policies in two steps. First, it programmatically synthesizes a large dataset of demonstrations from an MPC expert policy with privileged state information using our simulator, FiGS. Then, it distills these demonstrations into a policy deployed on the drone.
\vspace{-4mm}
\begin{figure}[h]
  \centering
  \includegraphics[width=0.90\columnwidth]{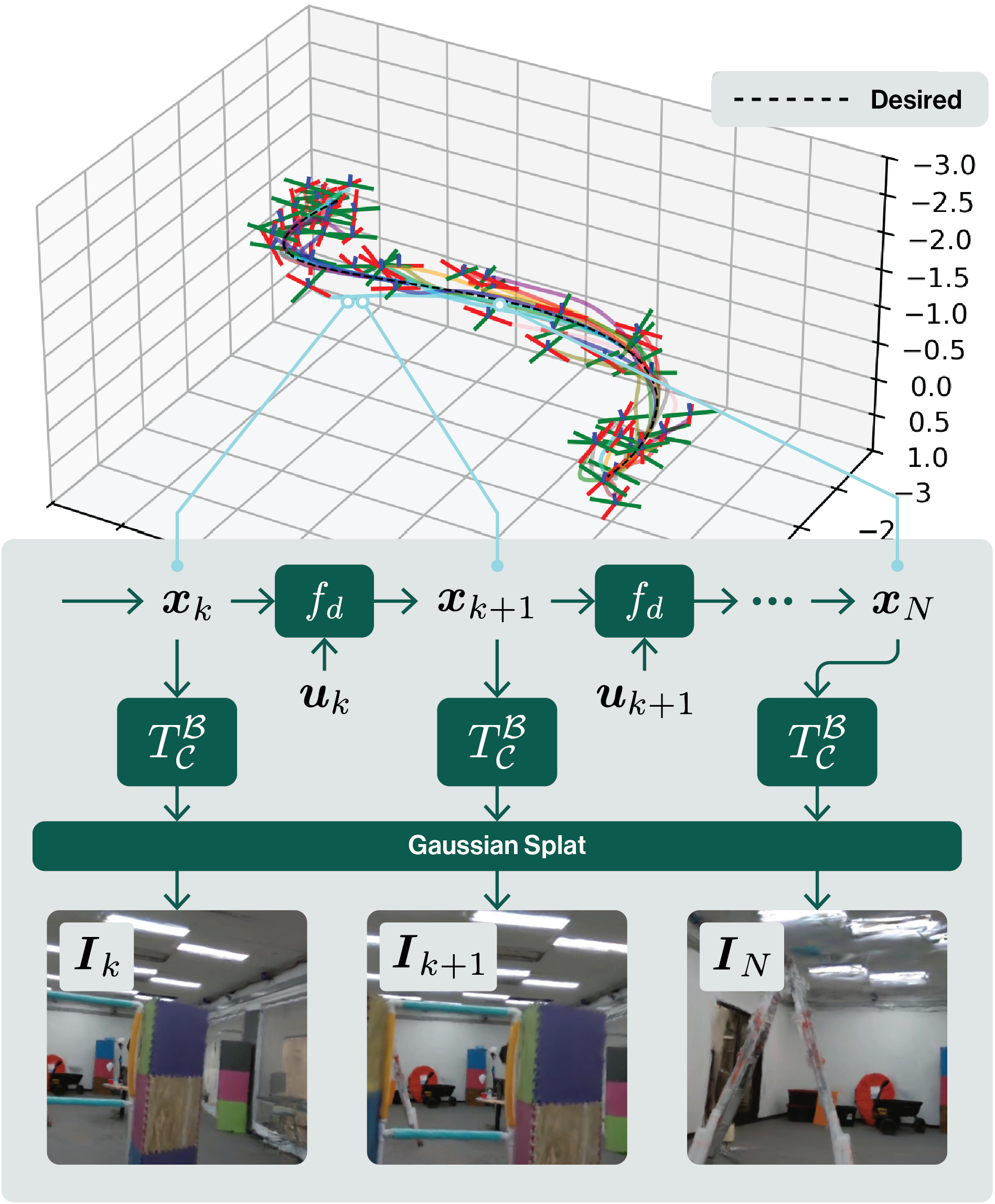}
\vspace{-2mm}
  \caption{Dynamic rollout of 50 data samples. At each time step, the update function $f_d$ simulates the solution from the MPC expert, while the transform $T_\mathcal{C}^\mathcal{B}$ is used to extract the corresponding camera image $I$ from the GSplat. }
  \label{fig:samples}
\end{figure}
\vspace{-2mm}

Many drone navigation frameworks are designed around a desired trajectory, whether to encode complex paths \cite{mellinger2011minimum,tagliabue2022efficient,tagliabue2024tube}, race courses over a sequence of gates \cite{kaufmann2023champion,gelesRSS2024demonstrating}, or even optimal approaches for perching \cite{thomas2015planning}. Given the complexity and variety of motion planning objectives, this abstraction facilitates a decoupled approach where high-level goals can be achieved by a higher-level task planner that generates a desired trajectory for a low-level navigation policy to execute. For instance, if the goal is obstacle avoidance, one could use the already existing GSplat to generate collision-free waypoints \cite{chen2024splat} that could then be turned into a desired trajectory.

In this work we are interested in the ability to navigate tight spaces and so we handpick a sequence of waypoints that intentionally guides the drone near or through obstacles. From these we use \cite{mellinger2011minimum} to compute a dynamically feasible spline which we then sample at our desired control frequency $\nu_{\text{ctl}}$ to extract an $N_d$-step state and input desired trajectory $(\mathbf{X}^{d},\mathbf{U}^{d})$, parametrized by $\bm{\theta}$. We can then apply one of a variety of trajectory optimization techniques, such as the one-shot sampling methods in \cite{tagliabue2022efficient} or even an iterative form of DAgger \cite{ross2011reduction}, to guide a drone towards the desired trajectory in simulation. We opt for the simplest approach, domain randomization, as described in Algo.~\ref{algo:domain_randomization} and illustrated in Fig.~\ref{fig:samples}. This leverages the strength of FiGS in quickly producing large volumes of photorealistic image data while leaving the door open to more sophisticated techniques.

Given a desired number of samples per time-step ($N_s$) and rollout duration ($t_s$), the demonstration dataset comprises of $N_s \cdot N_d$ dynamic rollout samples, each containing $\nu_{ctl} \cdot t_{s}$ time-steps of state ($\mathbf{X}^s$), input ($\mathbf{U}^s$) and image ($\bm{\mathcal{I}}^s$) data for a drone with parameters $\bm{\theta}_s$.
Each rollout begins by sampling $\bm{\theta}_s$ and $\bm{x}^s_0$ from a uniform distribution parametrized by ($\bm{\theta}_\text{min},\bm{\theta}_\text{max},\Delta \bm{x}$). $\bm{\theta}_s$ is then passed to \texttt{GenerateDynamics} to instantiate the dynamics update function ($f_d$) encoding (\ref{eq:equations_of_motion}). This enables us to run \texttt{MPC}, the expert policy which uses privileged information to guide the drone toward $\mathbf{X}^d$ from $\bm{x}^s_0$ by solving
\vspace{-1mm}
\begin{equation}\label{eq:mpc}
\begin{split}
    &\min_{\bm{u}}\sum_{k=0}^{N-1} (\delta \bm{x}_k^TQ_k\delta \bm{x}_k + \delta \bm{u}_k^TR_k\delta \bm{u}_k) + \delta\bm{x}_N^TQ_N\delta \bm{x}_N \\
    &\text{s.t.} \quad \bm{x}^{s}_{k+1} = f_d(\bm{x}^{s}_k, \bm{u}^{s}_k), \quad g_c(\bm{u}^{s}_k) \leq \bm{0}
\end{split}.
\vspace{-1mm}
\end{equation}
in an $N$-step receding horizon manner, subject to dynamics update $f_d$ and the control limits constraint $g_c$. The stage-wise weights, $(Q_k,R_k)$, and the terminal weights, $Q_N$, are applied to the difference between the rollout and the closest segment of the desired trajectory, defined by $\delta \bm{x}_k = (\bm{x}^{s}_k - \bm{x}^{d}_k)$ and similarly for $\delta\bm{u}_k$. The resulting state trajectory is then fed into \texttt{GenerateImages} to render first-person-view (FPV) images ($\bm{\mathcal{I}}^s$) from $\mathcal{GS}_{\phi}$ using the body to camera transform ($T_\mathcal{C}^\mathcal{B}$).
\vspace{-2mm}
\begin{algorithm}[h]
\caption{FiGS Domain Randomization}
\begin{algorithmic}[1]
\Require $\mathcal{GS}_{\phi}$, $\mathbf{X}^{d}$, $\mathbf{U}^{d}$, $N_d$, $\bm{\theta}_\text{min}$, $\bm{\theta}_\text{max}$, $\Delta \bm{x}$, $N_s$, $t_s$, $T_\mathcal{C}^\mathcal{B}$
\State Initialize dataset $\mathcal{D} = \emptyset$
\For{$i = 0$ to $N_d$}
    \For{$j = 0$ to $N_s$}
        \State $\bm{\theta}_s \!\sim\! U(\bm{\theta}_\text{min}, \bm{\theta}_\text{max})$, $\bm{x}^s_0 \!\sim\! (\bm{x}^{d}_i-\Delta\bm{x}, \bm{x}^{d}_i+\Delta\bm{x})$
        \State $f_d \gets \texttt{GenerateDynamics}(\bm{\theta}_s)$
        \State $\mathbf{X}^{s}, \mathbf{U}^{s} = \texttt{MPC}(\bm{x}^s_0, f_d,t_s,\mathbf{X}^{d},\mathbf{U}^{d})$
        \State $\bm{\mathcal{I}}^{s} = \texttt{GenerateImage}(\mathbf{X}^{s},T_{\mathcal{C}}^{\mathcal{B}},\mathcal{GS}_{\phi})$
        \State $\mathcal{D} \gets \mathcal{D} \cup \{(\mathbf{X}^{s}, \mathbf{U}^{s}, \bm{\mathcal{I}}^{s},\bm{\theta}_s)\}$
    \EndFor
\EndFor
\end{algorithmic}
\label{algo:domain_randomization}
\end{algorithm}
\vspace{-4mm}
\section{SV-Net Policy Architecture}
\label{sec:sv-net}
Our policy architecture, SV-Net, runs on an Orin Nano onboard the drone at 20 Hz. To output thrust and body rate commands, $\bm{u} = (f_{th}, \bm{\omega})$, the policy relies solely on onboard data: (i) images from the onboard camera and (ii) height, velocity, and orientation estimates ($p_{z}, \bm{v}_{\mathcal{W}}, \bm{q}_{\mathcal{BW}}$) provided by an Extended Kalman Filter (EKF), which fuses data from an IMU, a downward-facing time-of-flight sensor, and an optical flow sensor. These inexpensive, compact sensors are common on hobby-grade drones, providing state estimates that, while not pinpoint precise, are useful for control—especially since most height-sensitive applications occur over reasonably level surfaces. Notably, SV-Net performs better with $(p_z,\bm{v}_\mathcal{W})$, even when overflying obstacles, than without them. Beyond serving as direct inputs to SV-Net, these estimates, along with timestamps, are used to compute the history data:
\begin{equation}\label{eq:discrete_dynamics}
\begin{split}
    \delta t^{k-1} = t^k-t^{k-1}, \qquad
    \delta \bm{p}_{\mathcal{W}}^{k-1} = \delta t^{k-1} \cdot \bm{v}_\mathcal{W}^k, \\
    \delta \bm{v}_{\mathcal{W}}^{k-1} = \bm{v}_\mathcal{W}^k-\bm{v}_\mathcal{W}^{k-1}, \qquad 
    \delta \bm{q}_{\mathcal{BW}}^{k-1} = \bm{q}_{\mathcal{WB}}^{k} \cdot \bm{q}_{\mathcal{BW}}^{k-1}.
\end{split}
\end{equation}
We use $\bm{v}_\mathcal{W}$ to infer $\delta\bm{p}$ as the drone cannot observe its lateral position. For brevity, we use superscript time indices.
\begin{figure}[t]
  \centering
  \includegraphics[width=0.95\columnwidth]{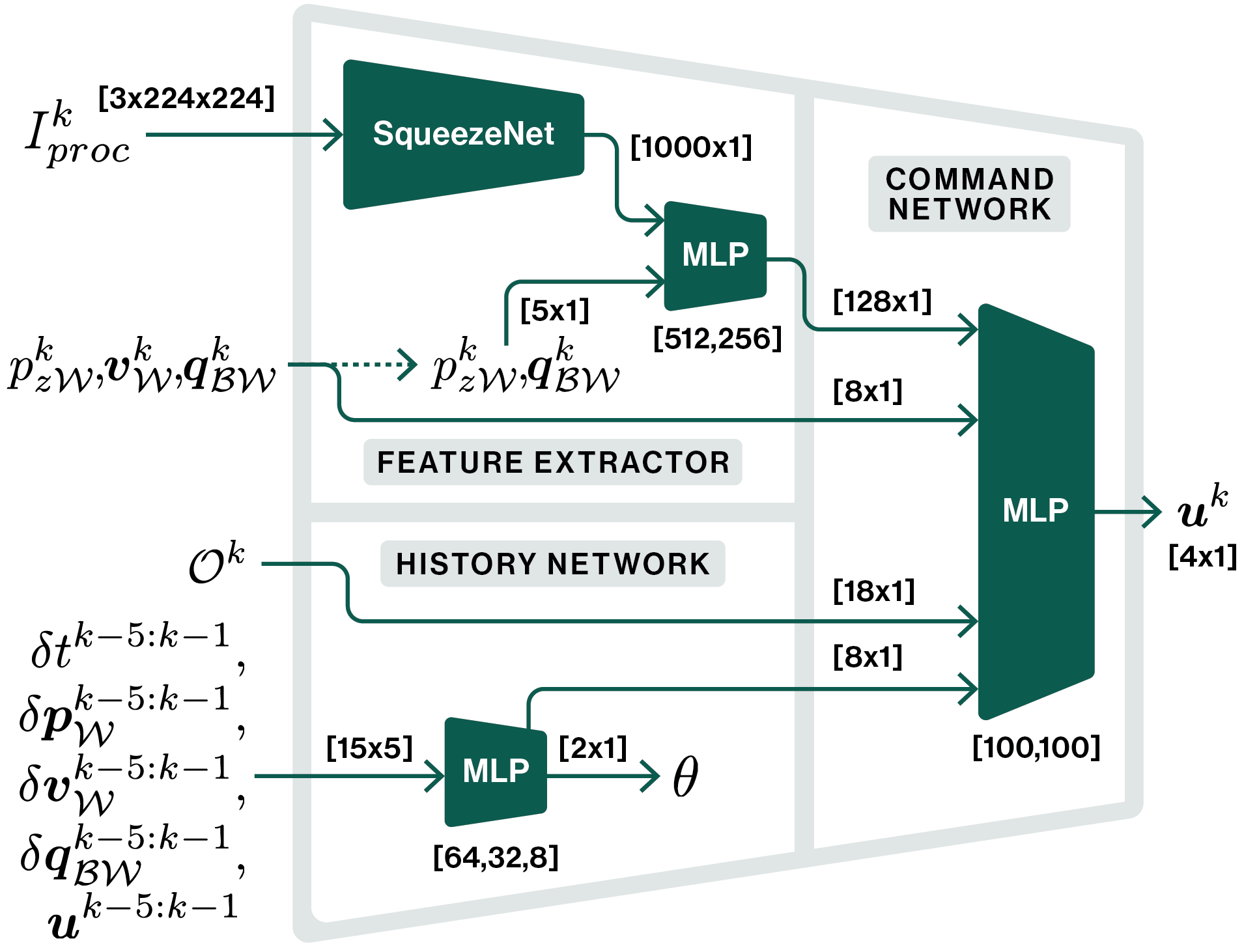}
  \vspace{-2mm}
  \caption{SV-Net consists of three components: a feature extractor that processes visual information from color images, a history network that uses an RMA technique to adapt to variations in dynamics through a history of observable states, and a command network that integrates the outputs of these components with observable states to generate body-rate commands.}
  \label{fig:network_diagram}
  \vspace{-4mm}
\end{figure}

SV-Net comprises three components: a feature extractor, a history network and a command network (Fig.~\ref{fig:network_diagram}). The architecture uses SqueezeNet \cite{iandola2016squeezenet} as a vision encoder, augmenting its output with estimated height and orientation before passing it through an MLP to create a pose-aware feature extractor. The history network, inspired by RMA, uses the sliding time-step window of history data to generate a latent vector encoding the evolving flight dynamics of the drone at that instant. The policy ingests the latent vector to adapt its output to current flight conditions. The command network combines the outputs of the feature extractor and history network with the observable states and an objective vector, $\mathcal{O}^k$, which encodes the change in position, initial and final velocity, initial and final orientation (quaternion), and total trajectory time. We use this to facilitate training and deployment across different trajectories when a single SV-Net is encoded with multiple trajectories (Section \ref{ssec:skilltest}).
\begin{figure*}[t]
    \centering
    \includegraphics[width=0.95\textwidth]{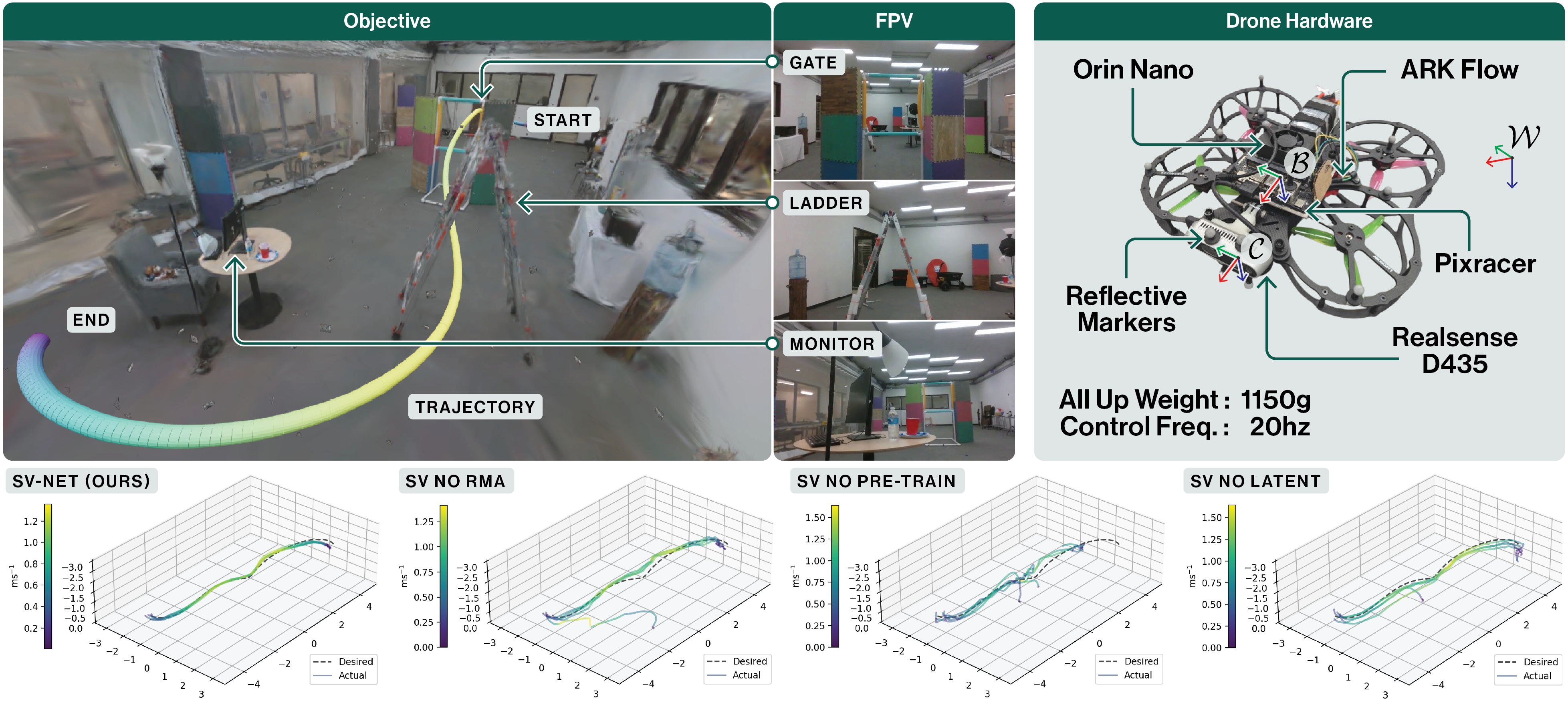}
    \vspace{-4mm}
    \caption{Clockwise from top left: 1) Desired trajectory in the scene's GSplat with corresponding real-world First-Person-View (FPV) of key objects. 2) Drone hardware and frames $(\mathcal{W},\mathcal{B},\mathcal{C})$. We use an Orin Nano and PixRacer Pro for control, while sensing is handled by the PixRacer's IMU, an ARK Flow sensor, and the D435's monocular camera. Motion capture markers provide ground truth. 3) 3D position and velocity performance of the policies in Section \ref{ssec:architecture_experiments}.}
    \label{fig:firstflights}
    \vspace{-4mm}
\end{figure*}

We train SV-Net on the demonstration dataset $\mathcal{D}$ in two stages. In the first stage, we train the history network to estimate $\bm{\theta}_s$ given history data extracted from $\mathbf{X}^s,\mathbf{U}^s$ through (\ref{eq:discrete_dynamics}). Once trained, the history network is frozen and the remaining components of the policy are trained end-to-end (including the SqueezeNet image encoder) to predict the body rate commands ($\mathbf{U}^s$) given the observable states within $\mathbf{X}^s$ and the images ($\bm{\mathcal{I}}^s$). In hardware testing, we find the best performance is achieved by using the second-to-last layer of the history network as input to the command network MLP, rather than the explicit estimate of $\bm{\theta}$.
\vspace{-3mm}

\subsection{Analysis of RMA Module}  \label{ssec:onion}
A property of (\ref{eq:equations_of_motion}) is that if we allow the drone parameters, $k_{th}$ and $m$, to be variables that can be adjusted online, we can use them to compensate for a wide range of model inaccuracies that are not limited to the thrust and weight of the drone. For simplicity, let $c=\frac{k_{th}}{m}$. Given an additional force vector $\bm{f}_{add}$ in the world frame, to account not only for model inaccuracies within $c$ but also for external forces such as aerodynamic drag and ground effect, we can compute an equivalent $\hat{c}$ in an augmented form of the velocity equation in (\ref{eq:equations_of_motion}),
\vspace{-1mm}
\begin{equation}\label{eq:augmented_velocity}
\begin{split}
g\bm{z}_\mathcal{W} - \hat{c}f_{th} \bm{z}_\mathcal{B} &= g\bm{z}_\mathcal{W} - cf_{th}\bm{z}_\mathcal{B} + \bm{f}_{add}
\end{split}.
\end{equation}
Then, taking the least-squares estimate of $\hat{c}$, we get:
\begin{equation}\label{eq:c_optimization}
\begin{aligned}
    \min_{\hat{c}} ||(\hat{c}-c)f_{th}\bm{z}_\mathcal{B}+\bm{f}_{add}||^2 \; \Rightarrow \; \hat{c} &\approx c - \frac{\bm{z}_\mathcal{B}^T\bm{f}_{add}}{f_{th}}.
\end{aligned}
\end{equation}
As evidenced in (\ref{eq:c_optimization}), the capacity of $\hat{c}$ to accurately approximate additional forces hinges on near-collinearity of $\bm{z}_{\mathcal{B}}$ and $\bm{f}_{add}$. This constraint is acceptable for most drone applications. For instance: (i) by definition, thrust-related additional forces align with the $\bm{z}_{\mathcal{B}}$ axis, (ii) much of the drone's operational envelope is near-hover, where $\bm{z}_{\mathcal{B}}$ aligns closely with primarily vertical forces, such as those due to changes in mass and ground effect, and (iii) at higher speeds, aerodynamic drag aligns with $\bm{z}_{\mathcal{B}}$, as the motor thrust vector must follow the flight direction. Hence, the RMA module can account for variations in flight dynamics the drone encounters during flight. 

\vspace{-2mm}
\section{Experiments} \label{sec:experiments}
In this section, we evaluate our SOUS VIDE policies across three fronts: efficacy of the proposed policy architecture, robustness to dynamic and visual disturbances, and generalization to novel scenarios. We demonstrate that the SV-Net policy, equipped with the RMA module, achieves state-of-the-art performance in zero-shot sim-to-real transfer. We emphasize that in all experiments, the policy does not observe the lateral position $(p_x,p_y)$. However, it does observe $p_z$ through the onboard time-of-flight sensor input.

We perform all experiments using a quadrotor drone equipped with a PixRacer Pro for low-level body-rate tracking control and an Orin Nano for policy execution,  as shown in Fig.~\ref{fig:firstflights}. The onboard sensing suite consists of an IMU, an ARK Flow sensor, and a monocular camera, with the first two fused via an EKF. The motion capture markers visible in our images and videos are used for diagnostics, enabling trajectory plotting in comparison to the ground truth.

To evaluate performance, we consider four key metrics. \textit{Completion}: Categorizes trajectories along a discrete spectrum--- (\cmark\cmark) indicates a fully successful position and orientation tracking with no collisions, (\cmark) allows for minor collisions with successful recovery, ($\bm{\sim}$) signifies completion of the position component but not the orientation, (\xmark) denotes failure due to an unrecovered collision, and (\xmark\xmark) corresponds to failure due to drifting off-course. \textit{Collision Rate (CR)}: Quantifies the number of collisions per meter traveled. \textit{Trajectory Tracking Error (TTE)}: Measures the $\ell_2$-norm of the position error relative to the closest point in the desired trajectory. Finally, \textit{Proximity Percentile (PP)}: Represents the fraction of the trajectory that remains within 30 cm of the intended path. Together, these metrics provide a comprehensive evaluation of trajectory accuracy, robustness, and recovery behavior.
\vspace{-2mm}

\subsection{Policy Architecture Ablations}\label{ssec:architecture_experiments}
We evaluated our main policy versus three ablations on a 15-second trajectory that guides the drone through a gate and under a ladder before ending facing a monitor. The desired trajectory is visualized in the GSplat in Fig.~\ref{fig:firstflights}, along with flights from each policy ablation.  All policies were trained on the same expert MPC dataset (180k observation-action pairs) using PyTorch, the Adam optimizer (learning rate 1e-4), for approximately 12 hours on a desktop machine (i9-13900K, RTX 4090, 64GB RAM). The policy ablations are:
\begin{itemize}
    \item \textbf{SV-Net}: Our proposed architecture with a locked pre-trained RMA network and 2nd-to-last layer latent code to the command network.
    \item \textbf{SV no RMA}: A minimal variant comprising only the feature extractor and command network. This serves as our approximation of a zero-shot transfer counterpart to the few-shot transfer described in \cite{tagliabue2024tube}.
    \item \textbf{SV no pre-train}: A variant of SV-Net that skips the RMA network pre-training and goes directly to training the entire network (with the history network unlocked).
    \item \textbf{SV no latent}: Same as SV-Net, but uses the RMA's explicit estimate of $\bm{\theta}$ instead of the 2nd-to-last layer.
\end{itemize}

\begin{table}[t]
\vspace{3mm}
\resizebox{\columnwidth}{!}{%
\begin{tabular}{@{}lcccccccc@{}}
        \toprule
        \multirow{2}{*}{\textbf{\begin{tabular}[c]{@{}l@{}}Ablation\\ Experiments\end{tabular}}} &
        \multicolumn{5}{c}{\textbf{Completion}} &
        \multirow{2}{*}{\begin{tabular}[c]{@{}c@{}}\textbf{CR}\\ (c/m)\end{tabular}} &
        \multirow{2}{*}{\begin{tabular}[c]{@{}c@{}}\textbf{TTE}\\ (m)\end{tabular}} &
        \multirow{2}{*}{\begin{tabular}[c]{@{}c@{}}\textbf{PP}\\ (\%)\end{tabular}} \\ \cmidrule(lr){2-6}
                & \textbf{\cmark\cmark} & \textbf{\cmark} & \textbf{$\bm{\sim}$} & \textbf{\xmark} & \textbf{\xmark\xmark} &       &       &      \\ \midrule
        SV-Net (ours)   & \textbf{5/5} &  -  &  -  &  -  &  -  & \textbf{0.00} & \textbf{0.17} & \textbf{96.0} \\
        SV no RMA       & 3/5 &  -  & 1/5 &  -  & 1/5 & 0.00 & 0.65 & 37.7 \\
        SV no pre-train &  -  &  -  & 1/5 & 4/5 &  -  & 0.10 & 0.39 & 51.4 \\
        SV no latent    & 1/5 & 2/5 & 1/5 & 1/5 &  -  & 0.05 & 0.43 & 42.0 \\
        \bottomrule
    \end{tabular}%
    }
\caption{Table comparing performance of ablations of SV-Net}
\label{tab:network_table}
\vspace{-6mm}
\end{table}
\begin{figure}[t]
  \centering
  \includegraphics[width=\columnwidth]{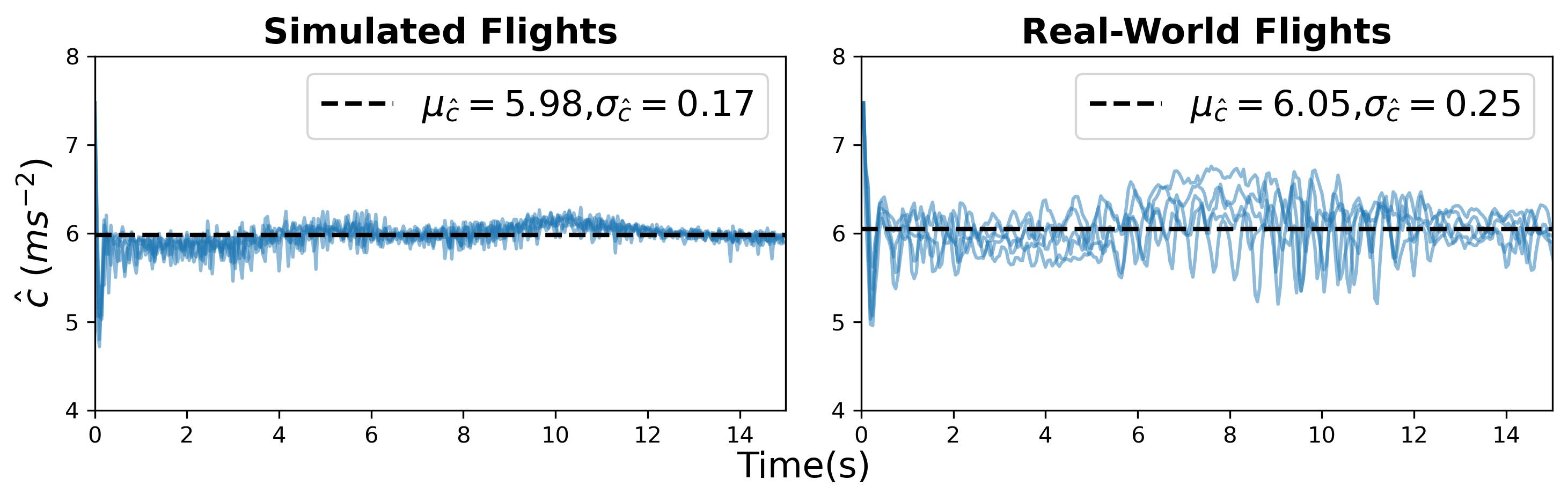}
  \vspace{-8mm}
  \caption{SV-Net history network's estimate of $\hat{c}$ with mean $\mu_{\hat{c}}$ overlaid for Section \ref{ssec:architecture_experiments} flights in simulation (left) and real-world (right).}
  \label{fig:ablation_c}
\vspace{-5mm}
\end{figure}

As shown in Table~\ref{tab:network_table}, \textbf{SV-Net} outperformed all other architectures, achieving a success rate of 100\% with no collisions, a TTE of 0.17m and a PP of 96\%, more than doubling the performance of \textbf{SV no RMA}.

To study the history network's performance, we acquired a ground truth estimate of $c=6.03$ by measuring the mass of the drone and recording the throttle command at hover. We found that when pre-trained (\textbf{SV-Net} and \textbf{SV no latent}), the RMA module maintained an estimated $\hat{c}$ value that stayed close to this in both simulation and real-world flights. SV-Net demonstrated the least deviation, with real-world $\mu_{\hat{c}}=6.05,\sigma_{\hat{c}}=0.25$ (illustrated in Fig.~\ref{fig:ablation_c}). In contrast, \textbf{SV no latent}'s estimate is more unstable with $\mu_{\hat{c}}=6.39,\sigma_{\hat{c}}=1.07$ across its five flights. We hypothesize that using the 2nd-to-last layer of the history network improves performance as its higher-dimensional latent code outweighs the minor information loss from skipping the final layer. Consequently, \textbf{SV no latent} suffers from a feedback loop, where poor estimates degrade policy performance, further amplifying estimation errors. We also note that \textbf{SV no pre-train}, which does have a history network but is instead trained directly on control commands, exhibits a highly unstable signal with ($\mu=-2.81,\sigma=9.56$).

Lastly, we observe that using larger datasets, while cheap to synthesize, offers little performance gain while increasing the training time.
\vspace{-2mm}

\subsection{Robustness Experiments} \label{ssec:robustness}
\begin{figure}[t]
  \centering
  \includegraphics[width=\columnwidth]{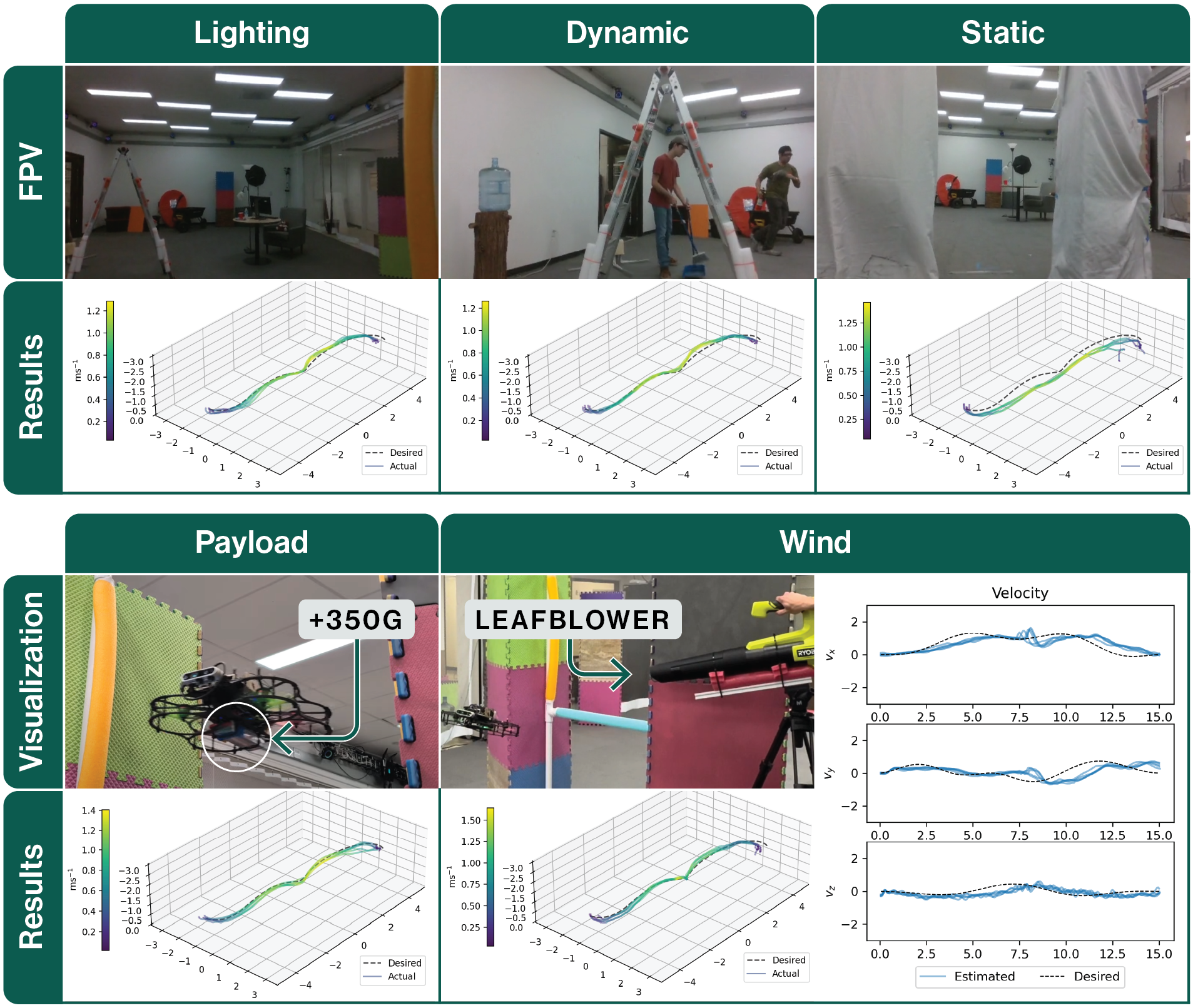}
  \vspace{-8mm}
  \caption{Visualization of disturbances and the corresponding position and velocity performance of SV-Net. Lighting: illumination reduced by 60\%, Dynamic: human activity in the scene, Static: key objects in training removed at runtime, Payload: 30\% increase in mass, Wind: 40 m/s gust from leaf blower. SV-Net maintains adequate performance in all cases.}
  \label{fig:robustvisual}
\end{figure}

Using the SV-Net result from Section \ref{ssec:architecture_experiments} as a baseline, we conduct five additional experiments, each introducing a distinct disturbance (illustrated in Fig.~\ref{fig:robustvisual}):
\begin{itemize}
    \item \textbf{Lighting}: Scene brightness was reduced to $40\%$ of original lumens.
    \item \textbf{Dynamic}: Four people actively moving within the field of view along the entire trajectory.
    \item \textbf{Static}: The gate, ladder, and monitor (present at train time) were removed at runtime while the pillars adjacent to the gate were occluded with white cloth.
    \item \textbf{Payload}: A rigid 350g payload ($30\%$ increase in drone weight) was attached below the center-of mass.
    \item \textbf{Wind}: The drone was exposed to a 40 m/s wind gust using a leaf blower.
\end{itemize}
\begin{table}[t]
\vspace{-2mm}
\resizebox{\columnwidth}{!}{%
\begin{tabular}{@{}lcccccccc@{}}
        \toprule
        \multirow{2}{*}{\textbf{\begin{tabular}[c]{@{}l@{}}Robustness\\ Experiments\end{tabular}}} &
        \multicolumn{5}{c}{\textbf{Completion}} &
        \multirow{2}{*}{\begin{tabular}[c]{@{}c@{}}\textbf{CR}\\ (c/m)\end{tabular}} &
        \multirow{2}{*}{\begin{tabular}[c]{@{}c@{}}\textbf{TTE}\\ (m)\end{tabular}} &
        \multirow{2}{*}{\begin{tabular}[c]{@{}c@{}}\textbf{PP}\\ (\%)\end{tabular}} \\ \cmidrule(lr){2-6}
                & \textbf{\cmark\cmark} & \textbf{\cmark} & \textbf{$\bm{\sim}$} & \textbf{\xmark} & \textbf{\xmark\xmark} &       &       &      \\ \midrule
        Baseline & 5/5 &  -  &  -  &  -  &  -  & 0.00 & 0.17 & 96.0 \\
        Lighting & 4/5 &  -  & 1/5 &  -  &  -  & 0.00 & 0.24 & 64.2 \\
        Dynamic  & 5/5 &  -  &  -  &  -  &  -  & 0.00 & 0.19 & 86.3 \\
        Static   &  -  &  -  & 4/5 &  -  & 1/5 & 0.06 & 0.49 & 25.3 \\
        Payload  & 5/5 &  -  &  -  &  -  &  -  & 0.00 & 0.20 & 80.1 \\
        Wind     & 5/5 &  -  &  -  &  -  &  -  & 0.00 & 0.20 & 81.7 \\ \bottomrule
    \end{tabular}%
    }
\caption{Table presenting SV-Net performance under disturbances.}
\label{tab:robustness_table}
\vspace{-9mm}
\end{table}
\begin{figure}[h]
  \centering
  \includegraphics[width=\columnwidth]{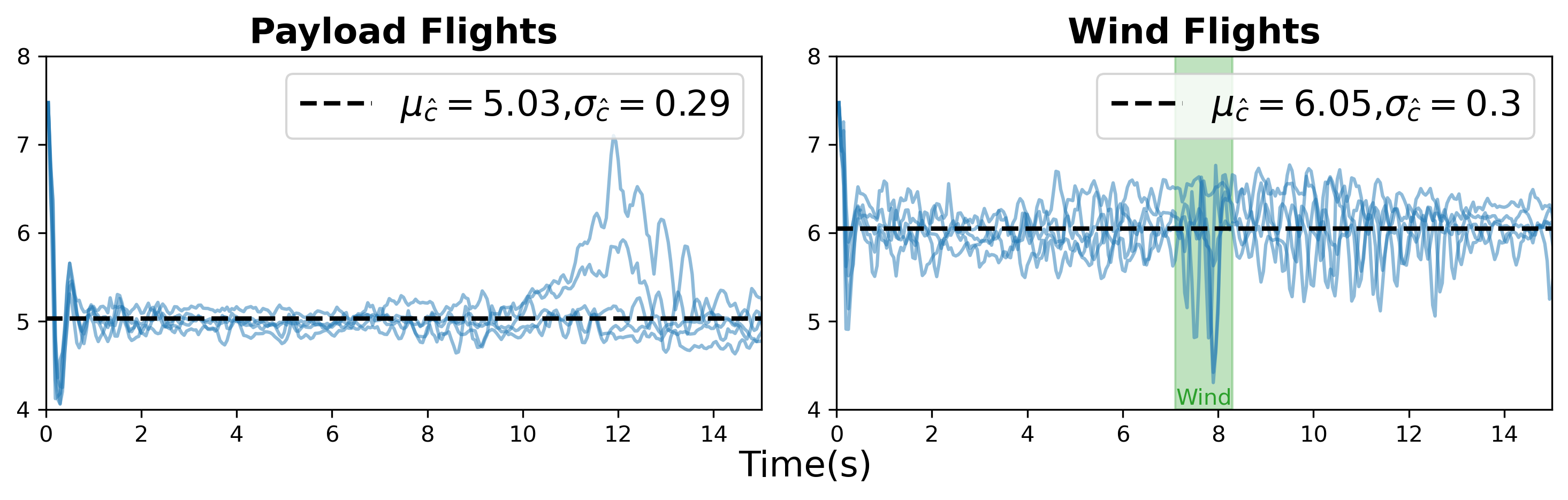}
  \vspace{-8mm}
  \caption{SV-Net estimate of $\hat{c}$ with mean $\mu_{\hat{c}}$ overlaid for Section \ref{ssec:robustness} flight: payload (left) and wind (right). Wind disturbance region highlighted in green.}
  \label{fig:robust_c}
\end{figure}

The results in Table~\ref{tab:robustness_table} show SV-Net consistently demonstrated resilience to dynamic disturbances, payload variations, and wind gusts, maintaining near-baseline performance with minimal impact across all metrics.

When lighting was degraded, the policy struggled to distinguish dark objects from the dim background, particularly near the end, where it drifted away from keeping the (black) monitor in frame. We also tested the policy with less than $40\%$ of the original lumens and the policy consistently failed by drifting off-course from the start location. While the policy handled dynamic scene changes with ease, static changes posed the greatest challenge: i) it underflew waypoints and experienced minor collisions with the occluded pillars, and (ii) it consistently flew through the space where the ladder rungs would have been. Despite these difficulties, the policy reliably tracked the overall trajectory shape, recovered from collisions, and successfully reached the final position in 4 out of 5 flights. These results suggest that the policy is able to retain essential scene information that would otherwise be lost in approaches relying on visual abstractions.

As shown in Fig.~\ref{fig:robust_c}, the RMA module maintains a stable \(\hat{c}\) under wind and payload disturbances, performing nearly identically to the baseline. In the wind disturbance flight, we see a downward spike in $\hat{c}$  that correlates to when the drone passes the leafblower (which is effecting a positive $\bm{f}_{add}$ on the drone). Interestingly, the estimated $\hat{c}$ during the payload flight is perceptibly different from the ground truth estimate updated with the additional mass ($c=4.62$). Given its overall trajectory performance, we believe the RMA module is in fact compensating for inaccuracies in the thrust model in (\ref{eq:equations_of_motion}), itself a simplified approximation of rotor dynamics.

\vspace{-2mm}
\subsection{Skill-Testing Experiments} 
\label{ssec:skilltest}
In our last set of experiments, we trained three different SV-Net policies, one for each of the following novel scenarios:
\begin{itemize}
    \item \textbf{Multi-Objective}: One policy was trained to execute three distinct trajectories within the same scene distinguished by unique objective inputs $O^k$. Two trajectories used identical positional splines but traversing in opposite directions, while the third followed a climbing orbit.
    \item \textbf{Extended Trajectory}: The drone navigated a trajectory that is double the length and duration of the trajectory in previous sections.
    \item \textbf{Cluttered Environment}: The policy was deployed close to the ground in a highly cluttered workshop with obstacles spaced as close as 1.0 m apart.
\end{itemize}
\vspace{-4mm}
\begin{figure}[h]
  \centering
  \includegraphics[width=\columnwidth]{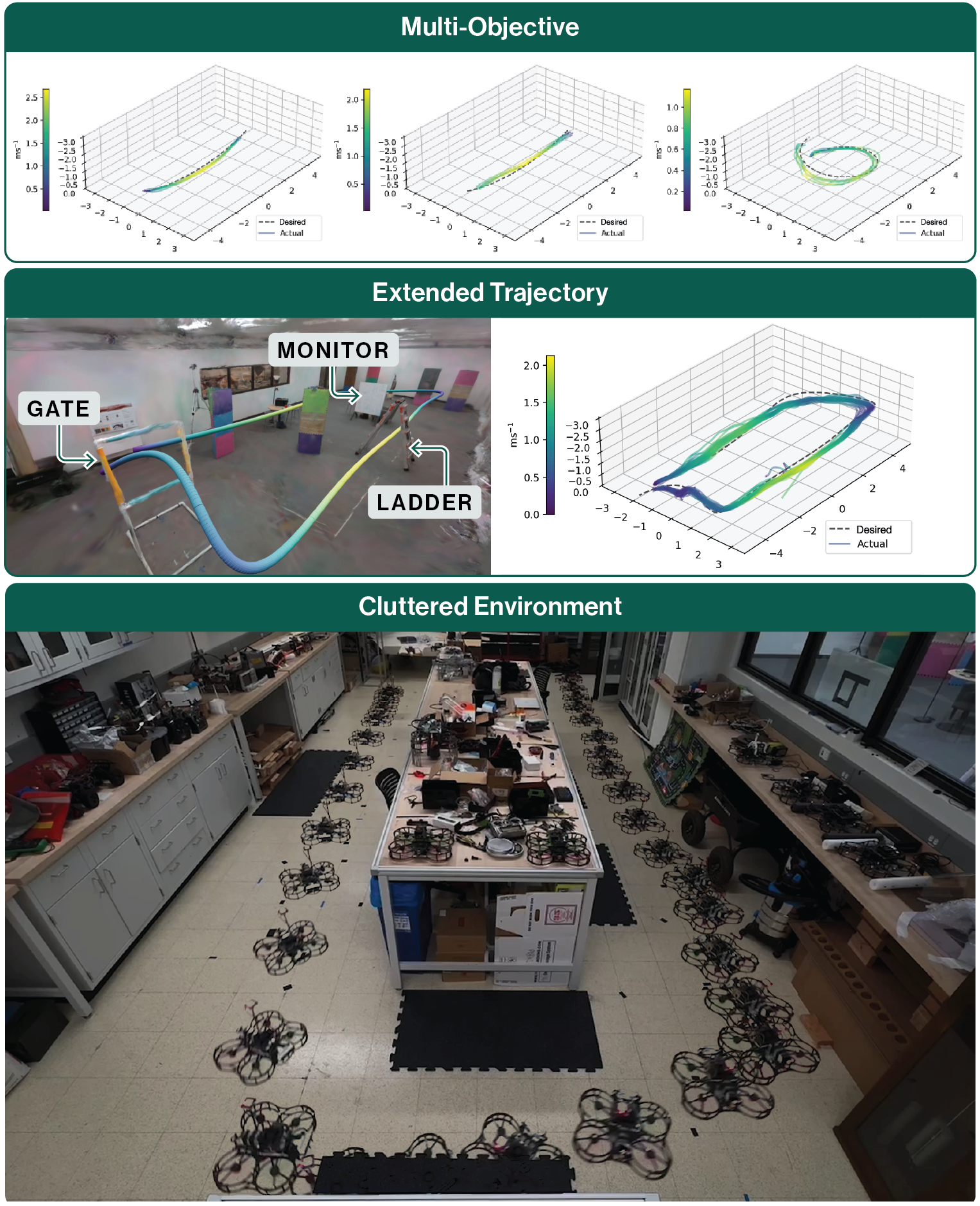}
    \vspace{-8mm}
  \caption{Position and velocity plots for the Multi-Objective (top) and Extended Trajectory (middle) experiments, with the latter’s desired trajectory in its GSplat. We also show a time-lapse of a Cluttered Environment flight (bottom).}
  \label{fig:skilltest}
\end{figure}

\begin{table}[h]
\vspace{3mm}
\resizebox{\columnwidth}{!}{%
\begin{tabular}{@{}lcccccccc@{}}
\toprule
\multirow{2}{*}{\textbf{\begin{tabular}[c]{@{}l@{}}Skill-Testing\\ Experiments\end{tabular}}} &
  \multicolumn{5}{c}{\textbf{Completion}} &
  \multirow{2}{*}{\begin{tabular}[c]{@{}c@{}}\textbf{CR}\\ (c/m)\end{tabular}} &
  \multirow{2}{*}{\begin{tabular}[c]{@{}c@{}}\textbf{TTE}\\ (m)\end{tabular}} &
  \multirow{2}{*}{\begin{tabular}[c]{@{}c@{}}\textbf{PP}\\ (\%)\end{tabular}} \\ \cmidrule(lr){2-6}
                & \textbf{\cmark\cmark} & \textbf{\cmark} & \textbf{$\bm{\sim}$} & \textbf{\xmark} & \textbf{\xmark\xmark} &       &       &      \\ \midrule
Multi-Objective       & 15/15 &   -   &   -   &   -   &   -   & 0.00 & 0.20 & 83.4 \\
Extended Trajectory   & 23/30 &  4/30 &   -   &  3/30 &   -   & 0.02 & 0.24 & 72.5 \\
Cluttered Environment & 13/15 &  1/15 &   -   &   -   &  1/15 & 0.01 &  n/a & n/a \\ \bottomrule
\end{tabular}%
}
\caption{Table presenting SV-Net performance over novel trajectories.}
\label{tab:skilltest_table}
\vspace{-6mm}
\end{table}
\begin{figure}[h]
  \centering
  \includegraphics[width=\columnwidth]{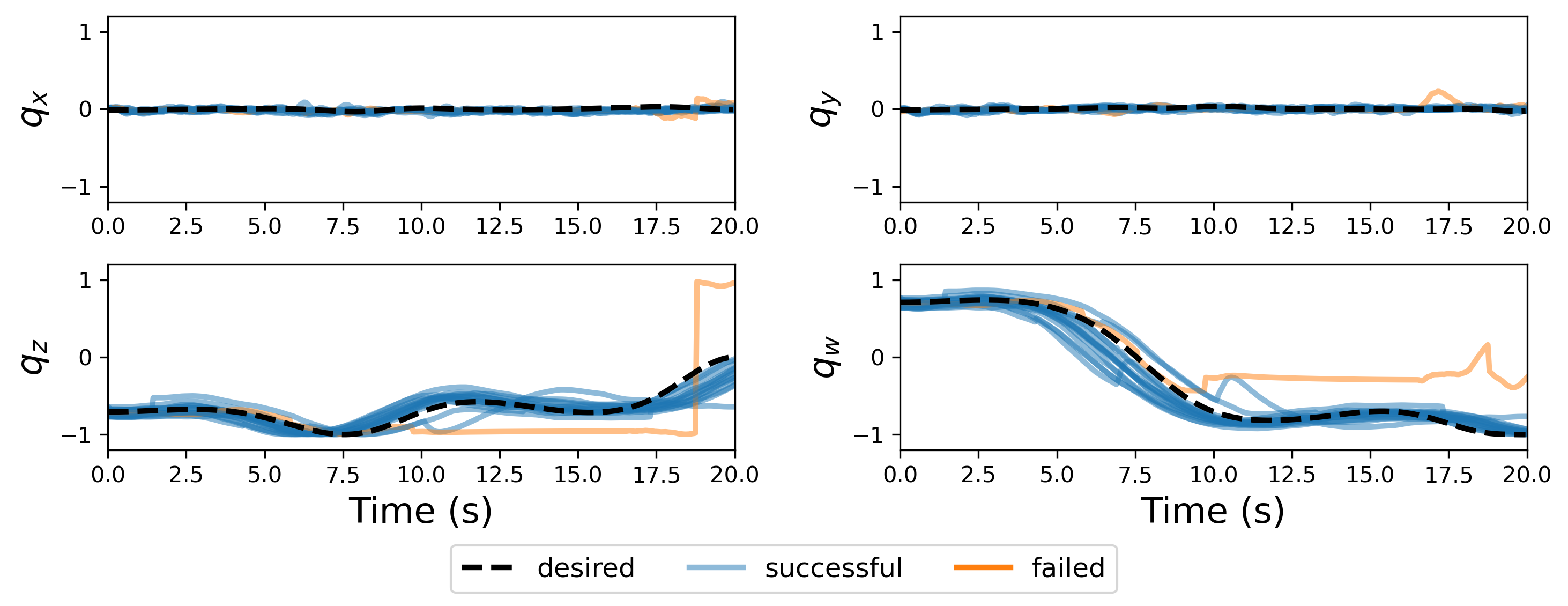}
    \vspace{-8mm}
  \caption{Quaternion rotation from the IMU during Cluttered Environment flights. The orange outlier trajectory is from the single failed flight.}
  \label{fig:c_skilltest}
\vspace{-5mm}
\end{figure}

Visualizations are shown in Fig.~\ref{fig:skilltest}, with results reported in Table~\ref{tab:skilltest_table}. The multi-objective policy had mixed success, indicating the need for more robust objective encodings in future work. Though it achieved a $100\%$ collision-free success rate, we observed significant degradation in one of the three tasks, where the drone consistently under-flew the desired trajectory and overshot its end-point In the extended trajectory, the policy performed comparably to the baseline in Section \ref{ssec:robustness}, with a low CR of 0.02 c/m, TTE of 0.24 m and a PP of 72.5\% over 30 attempts. Finally, in the cluttered environment, the policy achieved a $93.3\%$ success rate on a 20 s trajectory through a visually complex scene, demonstrating its robustness in real-world, unstructured environments. Exactly because it is an unstructured environment, there is no motion capture system available for measuring TTE and PP. Instead, we present a time-lapse (bottom of Fig.~\ref{fig:skilltest}) and the orientation reported by the onboard IMU (Fig.~\ref{fig:c_skilltest}).

\vspace{-2mm}
\section{Conclusions}
\label{sec:conclusions}
This work introduces the SOUS VIDE approach for training end-to-end visual drone navigation policies.  SOUS VIDE comprises the FiGS simulator based on a Gaussian Splat scene model, data generation from a simulated MPC expert, and distillation into a lightweight visuomotor policy architecture. By coupling high-fidelity visual data synthesis with online adaptation mechanisms, SOUS VIDE achieves zero-shot sim-to-real transfer, demonstrating robustness to variations in mass, thrust, lighting, and dynamic scene changes. Our experiments underscore the policy's ability to generalize across diverse scenarios, including complex and extended trajectories, with graceful degradation under extreme conditions. Notably, the integration of a streamlined adaptation module enables the policy to overcome limitations of prior visuomotor approaches, offering a computationally efficient yet effective solution for addressing model inaccuracies.  These findings highlight the potential of SOUS VIDE as a foundation for future advancements in autonomous drone navigation.

\textit{Limitations and Future Work:} While its robustness and versatility are evident, challenges such as inconsistent performance in multi-objective tasks suggest opportunities for improvement through more sophisticated objective encodings. SOUS VIDE has been used to train policies that are highly optimized for a single real-life environment. Future work will explore training policies with the same tools across multiple environments in FiGS to enable generalist skills, like general collision avoidance, and scene-agnostic navigation. We will also explore augmenting SOUS VIDE policies with semantic goal understanding, so goals can be given by a human operator in the form of natural language commands. Ultimately, this work paves the way for deploying learned visuomotor policies in real-world applications, bridging the gap between simulation and practical autonomy in drone operations.

\bibliographystyle{IEEEtran}
\bibliography{example}
\end{document}